\documentclass[twoside]{article}

\usepackage[accepted]{notaistats2023}

\usepackage[round]{natbib}

\usepackage[title]{appendix}
\usepackage{graphicx}
\usepackage{booktabs}
\usepackage{xcolor}
\usepackage{amssymb}

\newif\ifnotes
\notestrue %
\ifnotes
\newcommand{\mnote}[1]{\textcolor{magenta}{\textbf{Matt}: #1 }}
\newcommand{\tnote}[1]{\textcolor{blue}{\textbf{Tuan Anh}: #1 }}
\newcommand{\pnote}[1]{\textcolor{yellow}{\textbf{Pavel}: #1 }}
\newcommand{\bnote}[1]{\textcolor{green}{\textbf{Ben}: #1 }}
\newcommand{\cnote}[1]{\textcolor{green}{\textbf{Chris}: #1 }}
\newcommand{\rnote}[1]{\textcolor{green}{\textbf{Rif}: #1 }}
\else
\newcommand{\mnote}[1]{}
\newcommand{\tnote}[1]{}
\newcommand{\pnote}[1]{}
\newcommand{\bnote}[1]{}
\newcommand{\cnote}[1]{}
\newcommand{\rnote}[1]{}
\fi

\usepackage{hyperref}
\usepackage[capitalise]{cleveref}
\usepackage{booktabs}
\usepackage[colorinlistoftodos]{todonotes}

\crefname{appsection}{Appendix}{Appendices}

\begin{document}

\twocolumn[

\aistatstitle{ProbNeRF: Uncertainty-Aware Inference of 3D Shapes from 2D Images}

\aistatsauthor{Matthew D. Hoffman\footnotemark[1] \And Tuan Anh Le\footnotemark[1] \And Pavel Sountsov\footnotemark[1]}
\aistatsaddress{Google Research \And Google Research \And Google Research}

\aistatsauthor{Christopher Suter \And Ben Lee \And Vikash K. Mansinghka \And Rif A. Saurous}
\aistatsaddress{Google Research \And Google Research \And MIT, Google Research \And Google Research}

\runningauthor{Hoffman, Le, Sountsov, Suter, Lee, Mansinghka, Saurous}
]

\begin{abstract}
The problem of inferring object shape from a single 2D image is underconstrained. Prior knowledge about what objects are plausible can help, but even given such prior knowledge there may still be uncertainty about the shapes of occluded parts of objects. Recently, conditional neural radiance field (NeRF) models have been developed that can learn to infer good point estimates of 3D models from single 2D images. The problem of inferring uncertainty estimates for these models has received less attention. In this work, we propose probabilistic NeRF (ProbNeRF), a model and inference strategy for learning probabilistic generative models of 3D objects’ shapes and appearances, and for doing posterior inference to recover those properties from 2D images. ProbNeRF is trained as a variational autoencoder, but at test time we use Hamiltonian Monte Carlo (HMC) for inference. Given one or a few 2D images of an object (which may be partially occluded), ProbNeRF is able not only to accurately model the parts it sees, but also to propose realistic and diverse hypotheses about the parts it does not see. We show that key to the success of ProbNeRF are (i) a deterministic rendering scheme, (ii) an annealed-HMC strategy, (iii) a hypernetwork-based decoder architecture, and (iv) doing inference over a full set of NeRF weights, rather than just a low-dimensional code.

\footnotetext[1]{Equal contribution. \label{equal}}

\end{abstract}

\section{INTRODUCTION}

Neural radiance fields \citep[NeRFs;][]{mildenhall2020nerf} are remarkably good at estimating the 3D geometry of an object from 2D images of that object. A neural network (typically a modest-size multilayer perceptron) maps from 5D position-direction inputs to a 4D color-density output; this neural radiance field is plugged into a volumetric rendering equation \citep{blinn1982light} to obtain images of the field from various viewpoints, and trained to minimize the mean squared error in RGB space between the rendered images and the training images.

This procedure works well when the training images are taken from enough viewpoints to fully constrain the geometry of the scene or object being modeled. But it fails when only one or two images are available; one cannot infer 3D geometry from a single 2D image without prior knowledge about what sorts of shapes are plausible. To address this issue, various extensions of NeRF have incorporated implicit and explicit priors, yielding impressive one- or few-shot novel-view reconstructions and/or unconditional samples \citep{yu2021pixelnerf,kosiorek2021nerf,rebain2022lolnerf,rematas2021sharf,dupont2022data,jang2021codenerf,wang2021ibrnet,trevithick2021grf,chen2021mvsnerf}.

But even with a good shape prior, there may still be uncertainty about the shape and appearance of unseen parts of the object. Although existing approaches can infer reasonable point estimates from a single image, they generally fail to account for this uncertainty.

We propose probabilistic NeRF (ProbNeRF), a system for learning priors on NeRF representations of 3D objects and for doing inference on those representations. At a high level, ProbNeRF is trained using the variational autoencoder \citep{kingma2014auto, rezende2014stochastic} framework, using amortized variational inference to speed up training. At test time, we use Hamiltonian Monte Carlo \citep{neal2011mcmc} to sample from the posterior over NeRFs that are consistent with a set of input views.
Several technical contributions proved necessary to achieving high-fidelity reconstruction and robust shape uncertainty with this design:

\begin{figure*}[!t]
    \centering
    \includegraphics[width=0.65\linewidth]{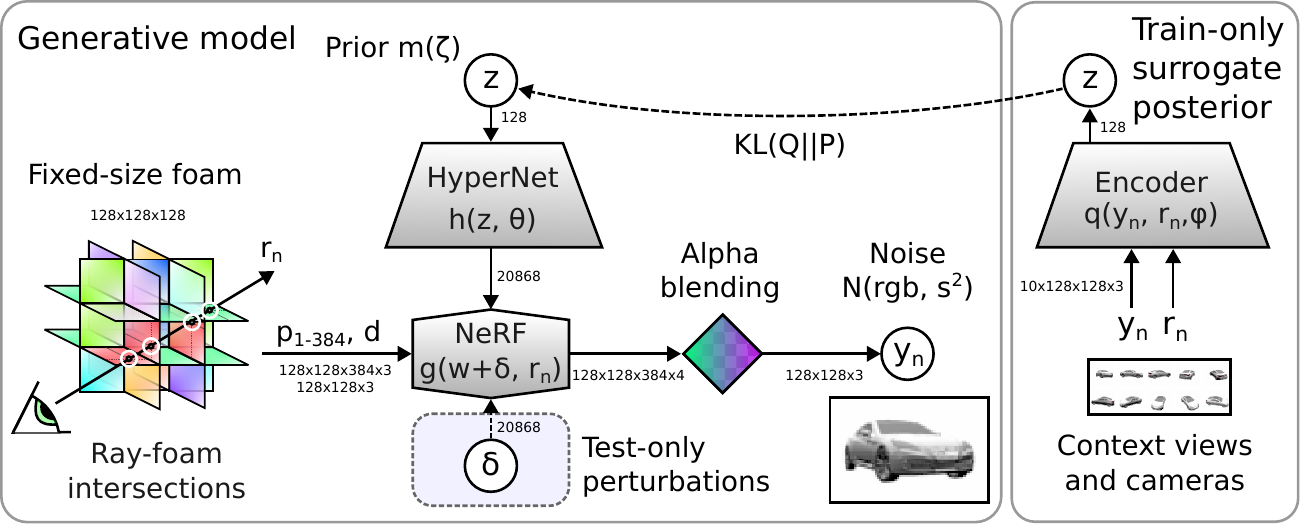}
    \caption{High-level diagram of the ProbNeRF generative process and training procedure. See \Cref{sec:method} for details.}
    \label{fig:diagram}
\end{figure*}

\begin{itemize}
    \item Instead of the Monte Carlo rendering strategy employed by \citet{mildenhall2020nerf} and most subsequent work, we use an exact renderer based on the work of \citep{chen2022mobilenerf}. HMC works well with this renderer, but rejects nearly all of its proposals with the standard renderer.
    \item We employ a temperature-annealing strategy in our HMC sampler to make it more robust to isolated modes that arise from the non-log-concave likelihood.
    \item We employ a two-stage hypernetwork-based decoder, rather than a single-network strategy such as latent concatenation. This design lets us represent each object using a relatively small NeRF, which dramatically reduces per-pixel rendering costs (and therefore the cost of iterative test-time inference).
    \item In addition to a low-dimensional latent code, we treat the raw weights of each object's NeRF representation as random variables to be inferred. This eliminates the latent-code bottleneck from our model, allowing high-fidelity reconstruction of novel objects.
\end{itemize}

\section{METHOD}
\label{sec:method}

\begin{figure*}[!t]
    \centering
    \includegraphics[width=\linewidth]{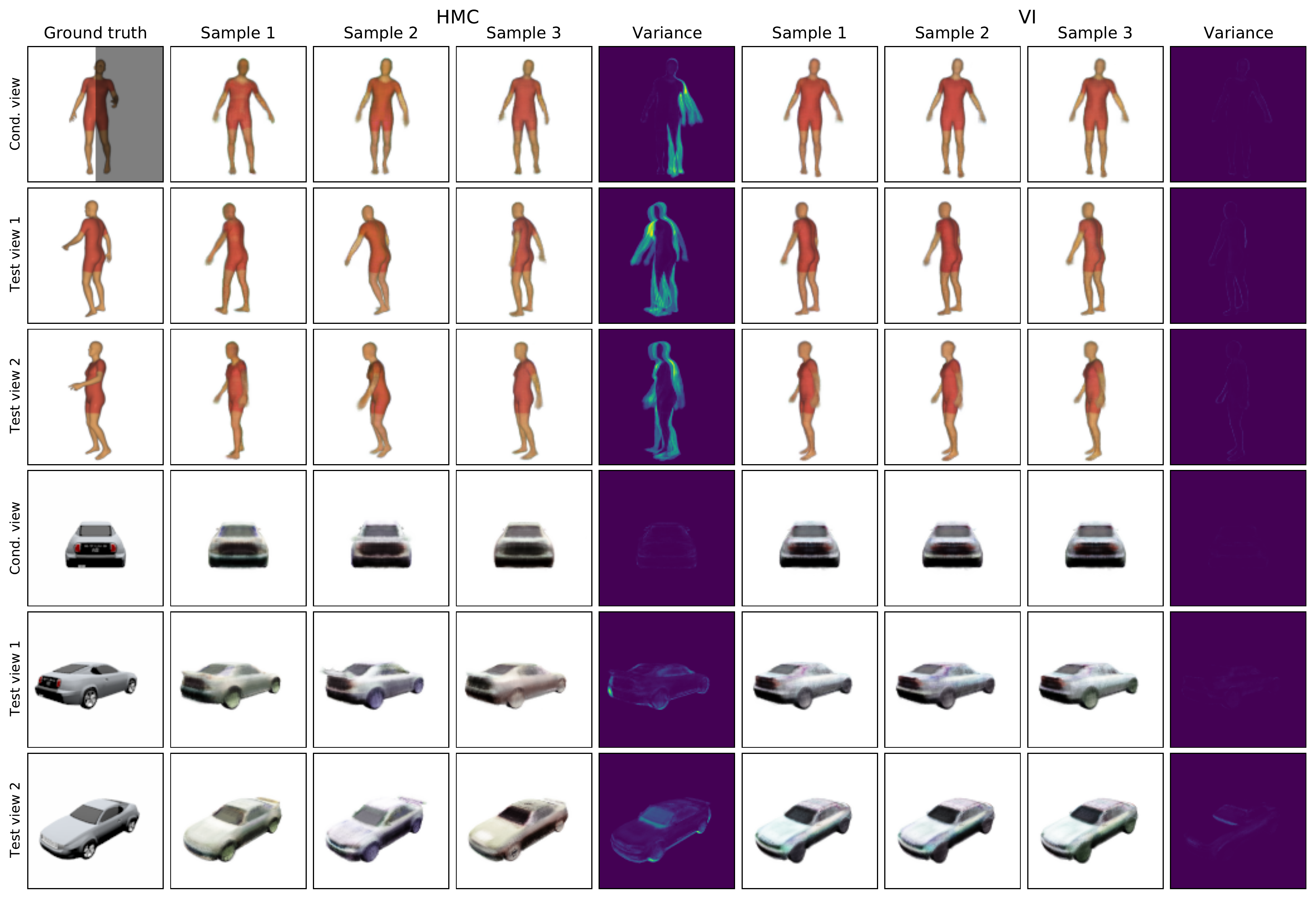}
    \caption{Conditioned on either the left half a view of a GHUM body (top left) or a back view of an SRN car (1st column, 4th row), HMC produces samples (columns 2--4) that are realistic, consistent with the conditioned-on view, and diverse as shown by the per-pixel variance (column 5). VI produces realistic and consistent samples (columns 6--8), but they have almost no diversity (last column). Renderings from novel views (rows 2--3/5--6) highlight the diversity of the HMC samples, in e.g. the poses of the left arm and the left leg or the variation of the car's shape, spoiler and color.}
    \label{fig:ghum_and_srncars_samples_and_variance}
\end{figure*}

In this section, we describe the ProbNeRF generative process, training procedure, and test-time inference procedure, as well as the neural architectures that implement them.

\subsection{Generative Process}

Let $f_w(x, v)$ be a function that, given some neural network weights $w$, a position $x\in\mathbb{R}^3$, and a viewing direction $v\in\mathbb{S}^2$, outputs a density $\sigma\in\mathbb{R}^+$ and an RGB color $c\in[0, 1]^3$. Let $g(w, r)$ be a rendering function that maps from a ray $r$ and the conditional field $f_w$ to a color $y\in[0, 1]^3$ by querying $f_w$ at various points along the ray $r$. (The renderer we use is defined in detail in \cref{sec:foam}.)

ProbNeRF assumes that, given a set of rays $r_{1:N}$ a set of pixels $y_{1:N}$ is generated by the following process: sample an abstract object code $z$ from a standard normal distribution pushed forward through an invertible RealNVP map $m$ \citep{dinh2017density}, run it through a hypernetwork to get a set of NeRF weights $w$, perturb those weights with low-variance Gaussian noise, render the resulting model, and add some pixelwise Gaussian noise. More formally,
\begin{align}
\label{eq:generative}
        \tilde z\sim\mathcal{N}(0, I);\quad
        z = m(\tilde z; \zeta);&\quad
        w = h(z; \theta);\quad
        \\
        \delta \sim\mathcal{N}(0, I);\quad
        \tilde w = w + \sqrt{\alpha}\delta;&\quad
        y_n\sim\mathcal{N}(g(\tilde w, r_n), s^2),
        \nonumber
\end{align}
where $m(\cdot; \zeta)$ is an invertible RealNVP \citep{dinh2017density} function with parameters $\zeta$, $z\in\mathbb{R}^K$ is a latent code that summarizes the object's shape and appearance, $h(z; \theta)$ is a hypernetwork with parameters $\theta$ that maps from codes $z$ to NeRF weights $w$, and $\alpha$ and $s^2$ are scalar variance parameters.
The generative process is summarized in \Cref{fig:diagram}.

This generative process is similar to the one assumed by \citet{kosiorek2021nerf,dupont2022data}: a latent code $z$ is sampled from a learned prior defined by a RealNVP, and used to index a learned family of NeRFs. There are two main differences. The first difference is architectural: we use a hypernetwork \citep{ha2017hypernetworks} to generate a full set of NeRF weights instead of concatenating the latent code $z$ to the input and activations\footnote{\citet{dupont2022data} frame their approach in terms of FiLM-style modulations \citep{perez2018film} rather than concatenation; we show in the supplement that their latent-shift strategy is equivalent to concatenating a latent code to the activations at each layer.}. This hypernetwork approach generalizes the latent-concatenation approach, and recent theoretical results \citep{galanti2020modularity} argue that hypernetworks should allow us to achieve a similar level of expressivity to the latent-concatenation strategy using a smaller architecture for $f$---intuitively, putting many parameters into a large, expressive hypernetwork makes it easier to learn a mapping to a compact function representation. This leads to large savings at both train and test time if we need to render many rays per object, since we can amortize the cost of an expensive mapping from $z$ to $w$ over hundreds or thousands of rays, each of which requires many function evaluations to render. For comparison, the NeRF architecture employed by \citet{dupont2022data} is an MLP with 15 layers of 512 hidden units, whereas in our experiments we get competitive results using a four-hidden-layer architecture with 64 hidden units---a cost savings of more than two orders of magnitude per function evaluation. Without this reduction in rendering costs, iterative MCMC methods for test-time inference would be impractical.

The second main difference between the ProbNeRF generative process and previous generative models of NeRFs is that we allow for small perturbations of the weights $w$. This is essentially a measure to address \emph{misspecification} \citep[cf. e.g.,][]{kleijn2012bernstein}; it ensures that our prior on NeRFs has positive support on the full range of functions $\{f_{\tilde w}\mid \tilde w\in\mathbb{R}^D\}$, rather than the much smaller manifold of functions $\{f_w\mid w = h(z; \theta) \textrm{ for some } z\in\mathbb{R}^K\}$. We choose the variance $\alpha=0.025^2$ on the weights to be small enough not to introduce noticeable artifacts, but large enough that the likelihood signal from a high-resolution image can overwhelm the prior preference to stay near the manifold defined by the mapping from $z$ to $w$. That way, even if the range of the hypernetwork does not include a parameter vector $w$ that accurately represents an object (for example, due to limited capacity or overfitting), the posterior $p(\tilde w\mid r, y)$ will still concentrate around a good set of parameters $\tilde w$ with more data. 

\subsection{Training Procedure}

We train ProbNeRF models using a variational autoencoder \citep{kingma2014auto, rezende2014stochastic} strategy, with a simplified generative process that omits the perturbation from $w$ to $\tilde w$:
\begin{equation}
\label{eq:generative-nodeltas}
    \begin{split}
        \tilde z\sim\mathcal{N}(0, I);\quad
        z = m(\tilde z;\zeta);&\quad
        w = h(z; \theta);
        \\y_n\sim\mathcal{N}(g(w, r_n),s^2)&.
    \end{split}
\end{equation}
We omit these perturbations at training time to force the model to learn hypernet parameters $\theta$ and RealNVP parameters $\zeta$ that can explain the training data well without relying on perturbations. The perturbations $\delta$ are intended to allow the model as an inference-time ``last resort'' to explain factors of variation that were not in the training set; at training time we do not want $\delta$ to explain away variations that could be explained using $z$, since the model lacks a mechanism to learn a meaningful prior on $\delta$.

To compute a variational approximation $q(z\mid y, r)$ to the posterior $p(z\mid y, r)$, we use a convolutional neural network \citep[CNN;][]{lecun1998convolutional} to map from each RGB image and camera matrix to a diagonal-covariance $K$-dimensional Gaussian potential, parameterized as locations $\mu_j$ and precisions $\tau_j$ for the $j$th image; these potentials are meant to approximate the influence of the likelihood function on the posterior \citep{johnson2016composing,sonderby2016ladder}. We combine these $J$ potentials with a learned ``prior'' potential parameterized by location $\mu_0$ and precisions $\tau_0$ via the Gaussian update formulas
\begin{equation}
\textstyle
    \hat\tau = \sum_{j=0}^J \tau_j;\quad
\hat\mu = \hat\tau^{-1}\sum_{j=0}^J\tau_j\mu_j
\end{equation}
and set $q(z_k\mid y, r) = \mathcal{N}(z_k; \hat\mu_k, \hat\tau_k^{-1})$. 

We train the encoder parameters $\phi$, the hypernet parameters $\theta$, and the RealNVP parameters $\zeta$ by maximizing the evidence lower bound (ELBO) using Adam \citep{kingma2015adam}:
\begin{equation}
\begin{split}
    \mathcal{L} &=\textstyle
    \mathbb{E}_q[\log p(y\mid z, r) - \log \frac{q(z\mid y, r)}{p(z)}]
    \\ &=\textstyle
    \mathbb{E}_q\left[\log p(y\mid z, r) - \log \frac{q(z\mid y, r)}{\mathcal{N}(m^{-1}(z); 0, I)|\frac{dm^{-1}}{dz}|}\right].
\end{split}
\end{equation}
We train on minibatches of 8 objects and 10 randomly selected images per object to give the encoder enough information to infer a good latent code $z$. The encoder sees all 10 images, but to reduce rendering costs we compute an unbiased estimate of the log-likelihood $\log p(y\mid z, r)$ from a random subsample of 1024 rays per object.

\begin{figure*}[!t]
    \centering
    \includegraphics[width=0.95\linewidth]{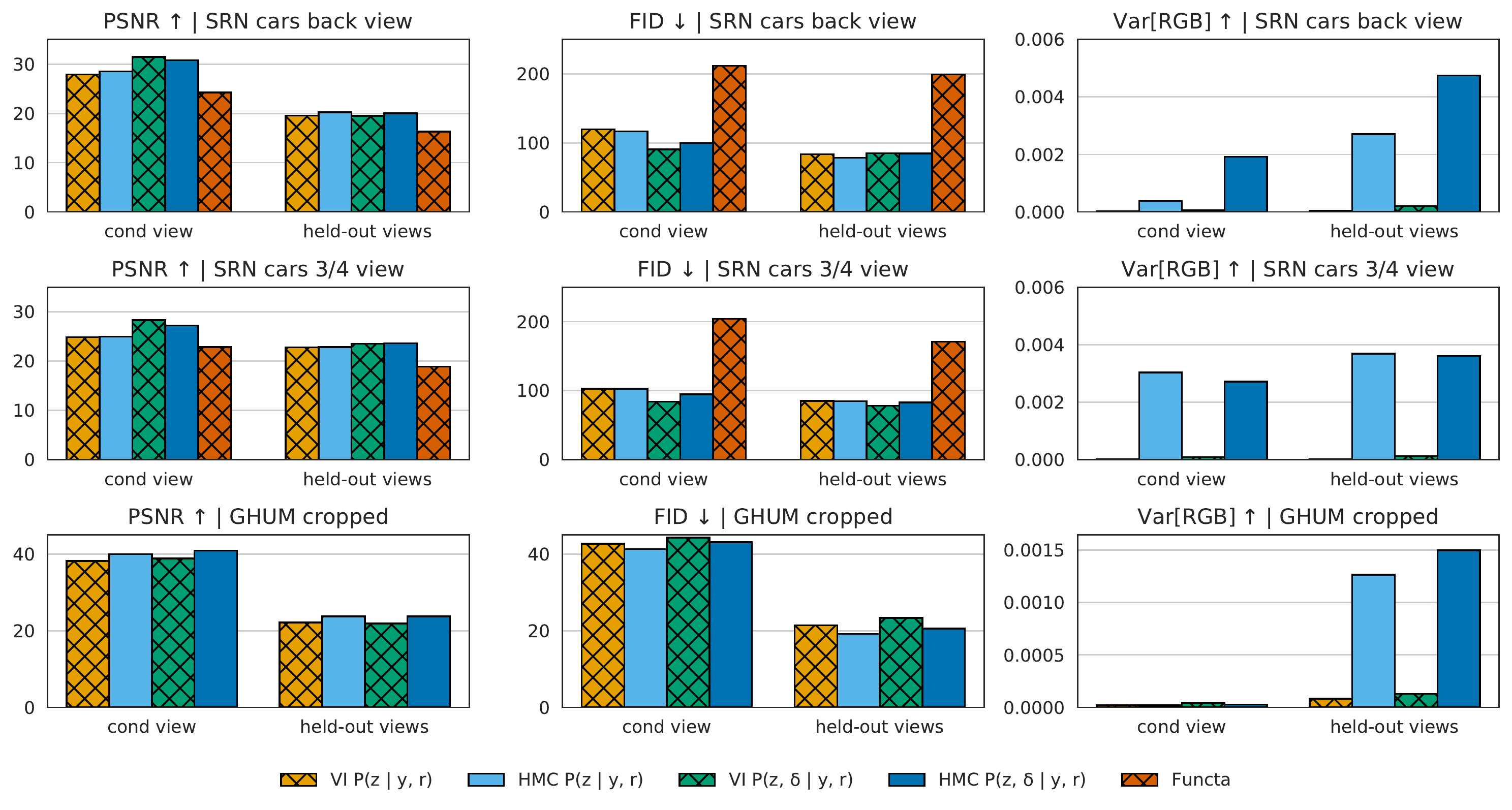}
    \caption{Posterior-predictive metrics for VI and HMC on the ProbNeRF model and MAP inference on Functa (where appropriate) on held-out data for SRN Cars and GHUM. For SRN cars, the top two rows condition on a uninformative back view and an informative 3/4 view respectively, and evaluate on 3 held-out views. For GHUM the bottom row is conditioned on the left-half crop of a front view, evaluated on 4 held-out views. We compare VI and HMC on full ProbNeRF ($p(\tilde z, \delta \mid y, r)$) and an ablation where the $\delta$ is omitted ($p(\tilde z \mid y, r, \delta=0)$). The first column shows reconstruction quality in PSNR. The middle column is FID score, measuring whether reconstructions look like test examples from the corresponding view(s). The last column shows the variance of RGB values in the reconstructed images, measuring posterior diversity.
    }
    \label{fig:ablation_evals}
\end{figure*}

\subsection{Architectures}
For each object's NeRF, we use two MLPs, each with two hidden layers of width 64. The first MLP maps from position to density; the second maps from position, view direction, and density to color. All positions and view directions are first transformed using a 10th-order sinusoidal encoding \citep{mildenhall2020nerf}. The number of parameters per object is 20,868, relatively few for a NeRF.

The RealNVP network that implements the mapping from $\tilde z$ to $z$ comprises two pairs of coupling layers. Each coupling layer is implemented as an MLP with one 512-unit hidden layer that shifts and rescales half of the variables conditioned on the other half; each pair of coupling layers updates a complementary set of variables. The variables are randomly permuted after each pair of coupling layers.

The hypernetwork that maps from the 128-dimensional code $z$ to the 20,868 NeRF weights is a two-layer 512-hidden-unit MLP. This mapping uses a similar number of FLOPs to rendering a few pixels (see \cref{sec:foam}).

The encoder network applies a 5-layer CNN to each image and a two-layer MLP to its camera-world matrix, then linearly maps the concatenated the image and camera activations to locations and log-scales for each image's Gaussian potential. (Full architectural details in supplement.)

All networks use ReLU nonlinearities.

\subsection{Test-Time Inference}
\label{sec:hmc}
The training procedure outlined above is able to learn a good RealNVP prior on codes and to reconstruct training-set examples accurately. However, we found that the trained encoder generalizes poorly to held-out examples---it is useful scaffolding for training the model, but fails to accurately reconstruct objects it was not trained on. Furthermore, variational inference is well known to underestimate posterior uncertainty \citep[e.g.,][]{yao2018yes}, and one of our primary goals is to capture uncertainty about object shape and appearance.

As an alternative, at inference time we turn to Hamiltonian Monte Carlo \citep[HMC;][]{neal2011mcmc}, a gradient-based Markov chain Monte Carlo (MCMC) method that uses momentum to mitigate poor conditioning of the target log-density function. Rather than sample in $z, \tilde w$ space, we use the noncentered parameterization and sample from $p(\tilde z, \delta\mid y, r)$ \citep{betancourt2015hamiltonian}, since the joint prior for $\tilde z$ and $\delta$ is a well-behaved spherical normal. (Note that only the prior's contribution to the posterior is simple; the likelihood still makes things difficult.)

HMC is a powerful MCMC algorithm, but it can still get trapped in isolated modes of the posterior. Running multiple chains in parallel can provide samples from multiple modes, but it may be that some chains find (but cannot escape from) modes that have negligible mass under the posterior. A conditioning problem also arises in inverse problems where some degrees of freedom are poorly constrained by the likelihood: as the level of observation noise decreases it becomes necessary to use a smaller step size, but the distance in the latent space between independent samples may stay almost constant \citep{langmore2021hamiltonian}.

To make our sampling procedure more robust to minor modes and poor conditioning, we use a temperature-annealing strategy \citep[e.g.,][]{kirkpatrick1983optimization,neal2001annealed}. Over the course of $T$ HMC iterations, we reduce the observation-noise scale $s$ logarithmically from a high initial value $s_0$ to a low final value $s_T$, with $s_t=s_0^{(T-t)/T}s_T^{t/T}$ (for a Gaussian likelihood, this is equivalent to annealing the ``temperature'' of the likelihood). That is, we start out targeting a distribution that is close to the prior, and gradually increase the influence of the likelihood until we are targeting the posterior. We also anneal the step size so that it is proportional to $s_t$. This procedure lets the sampler explore the latent space thoroughly at higher temperatures before settling into a state that achieves low reconstruction error. In \cref{sec:exp-annealing}, we show that this annealing procedure yields more-consistent results than running HMC at a low fixed temperature.

\subsection{Exact Rendering}\label{sec:foam}

NeRFs \citep{mildenhall2020nerf} generally employ a stochastic quadrature approximation of the rendering integral. Although this procedure is deterministic at test time, we have found empirically that its gradients are not reliable enough to use in HMC (see \cref{sec:exp-foam}).

While stochastic-gradient methods are robust to the noise from this procedure, standard HMC methods are not \citep{betancourt2015fundamental}. Stochastic-gradient HMC methods do exist \citep{ma2015complete}, but require omitting the Metropolis correction, which perturbs the stationary distribution unless one uses a small step size and/or can accurately estimate the (high-dimensional) covariance of the gradient noise.

Instead, we employ a simplified version of the renderer used in \citet{chen2022mobilenerf}.  We assume all density is concentrated in a ``foam'' consisting of the surfaces of a 128x128x128 lattice of cubes. Since there is no density inside the cubes, we can render a ray by enumerating all ray-cube intersection points, computing opacities and colors at each intersection, and alpha-compositing the result. This simplification avoids the need to map the latent code to grid vertices as in \citet{chen2022mobilenerf}. Rendering a ray requires at most $128\times3=384$ function evaluations (\emph{not} $128^3$). In \cref{sec:exp-foam} we show that this renderer works well with HMC, while HMC with the standard quadrature scheme cannot achieve high acceptance rates.

\section{EXPERIMENTS}

In this section, we qualitatively and quantitatively evaluate ProbNeRF's ability to generate realistic-looking conditional and unconditional samples, as measured by FID score \citep{heusel2017gans};
accurately reconstruct the views we condition on;
and, given a single image, generate diverse and plausible hypotheses about what an object looks like from other views.
We also apply a set of ablations to demonstrate the value of using an exact renderer when doing MCMC,
using temperature annealing in our HMC procedure,
and doing posterior inference over the raw NeRF weights as well as the latent code.

We evaluate ProbNeRF on two datasets: SRN Cars \citep{sitzmann2019scene}, and a set of renderings from the GHUM generative model \citep{xu2020ghum}.
We use the standard SRN Cars dataset with 2458 train cars and 704 test cars.
GHUM is a generative model fit on a dataset of 60,000 photo-realistic body scans that comprises of normally distributed latent codes corresponding to the facial expression, body pose, and body shape which are decoded into a 3D mesh which can be rendered given a camera pose.
To isolate the representation of uncertainty over body poses, we construct a dataset by sampling a code for body pose while setting the facial expression and body shape codes to zero.
We generate 2500 training examples, each containing 50 views from cameras pointing to the middle of the body with poses sampled uniformly from a fixed-radius circle around the body, parallel to the ground plane and elevated to the middle of the body.
We generate 100 test examples, each containing 50 views in a similar way except the camera poses are equally spaced points on the circle.
All images have resolution $128 \times 128$.
All models were trained for one million iterations with an observation noise scale $s=0.1$.

\subsection{Posterior Inference}
\label{sec:posterior_inference_probnerf}

We want ProbNeRF to accurately reconstruct the view it is conditioned on and produce realistic and diverse reconstructions of held out views.
To quantitatively assess this, we measure reconstruction quality using the PSNR metric \citep{wang2004image}, realism using the FID score \citep{heusel2017gans}, and diversity using average per-pixel variance.

We compare ProbNeRF with baseline methods along the inference and the modeling axes.
For inference, we compare ProbNeRF's HMC inference procedure with an iterative mean-field variational inference (VI) procedure fit using Adam and sticking-the-landing gradients \citep{roeder2017sticking}.
For modeling, we compare ProbNeRF with ``Functa'' \citep{dupont2022data} which proposes separately extracting a compressed representation---a functum---for each training example and doing probabilistic modeling and inference on top of the functa.

To compare between HMC and VI, we use the same trained ProbNeRF model and for each test example pick a conditioning view $(y, r)$ and a set of $H = 4$ held out views $\{(y_h', r_h')\}_{h = 1}^H$.
For both HMC and VI, we produce a set of $L$ samples $\{(\tilde z_\ell, \delta_\ell)\}_{\ell = 1}^L$ targeting the single view posterior $p(\tilde z, \delta | y, r)$ which are used to produce renderings from the conditioning view $y_\ell = g(\tilde w_\ell, r)$ and from the held-out view $y_h'^{\ell} = g(\tilde w_\ell, r_h')$. We chose a perturbation variance $\alpha$ of $0.025^2$ by rendering perturbed weights with different amounts of Gaussian noise and choosing the largest variance that did not produce significant artifacts.

In HMC, we obtain samples by running 8 independent chains with the annealing procedure described in \cref{sec:hmc} and taking the last 16 samples in each chain.
In all experiments, we run annealed HMC for $T=100$ steps with 100 leapfrog steps per HMC step, for a total of 10,000 gradient evaluations. The noise scale $s$ is annealed from $s_0=5$ to $s_T=0.1$ In VI, we approximate the posterior using an independent normal distribution for each element of the latent variable $(\tilde z, \delta)$ parameterized by the location $\mu$ and log scale $\log \sigma$.
We maximize the evidence lower bound with respect to $(\mu, \log \sigma)$ using 1500 gradient steps and generate $16$ samples using the final parameters.

We find that while both HMC and VI produce high-quality (high PSNR) and realistic (low FID) reconstructions, HMC produces significantly more-diverse held-out reconstructions (higher mean per-pixel variance) for both SRN Cars and GHUM (\cref{fig:ablation_evals}).
This is qualitatively illustrated in \cref{fig:ghum_and_srncars_samples_and_variance} where we show three samples for each HMC and VI rendered from the conditioning view and two held out views alongside the per-pixel variance shown as a heatmap (more HMC samples can be found in the supplement).
HMC samples show diversity in the pose of the left arm and the left leg in the GHUM example and in the variation of the spoiler, bumper and taillights in the SRN Cars example.
VI samples show almost no diversity.

To compare between ProbNeRF and Functa, we train a Functa model on SRN Cars using the open-sourced code to obtain a functaset of training-set modulations.
We fit the RealNVP prior used for our model instead of a neural spline flow \citep{durkan2019neural} or denoising diffusion probabilistic model \citep{ho2020denoising}, as done in Functa, since the prior-fitting code is not released.
We note that this may be a source of difference in our reproduction.
For each test example, we obtain a modulation code by performing a 1000-step gradient-based MAP search given the observed view $(y, r)$ as per equation 2 in \citep{dupont2022data}.
These modulation codes are then rendered for the conditioned view and the held-out views.

We find that Functa produces less-accurate, less-realistic, and less-diverse reconstructions on both conditioned on and held-out views than ProbNeRF (\cref{fig:ablation_evals}).

\subsection{Generative Modeling}

\begin{figure}[!t]
    \centering
    \includegraphics[width=\linewidth]{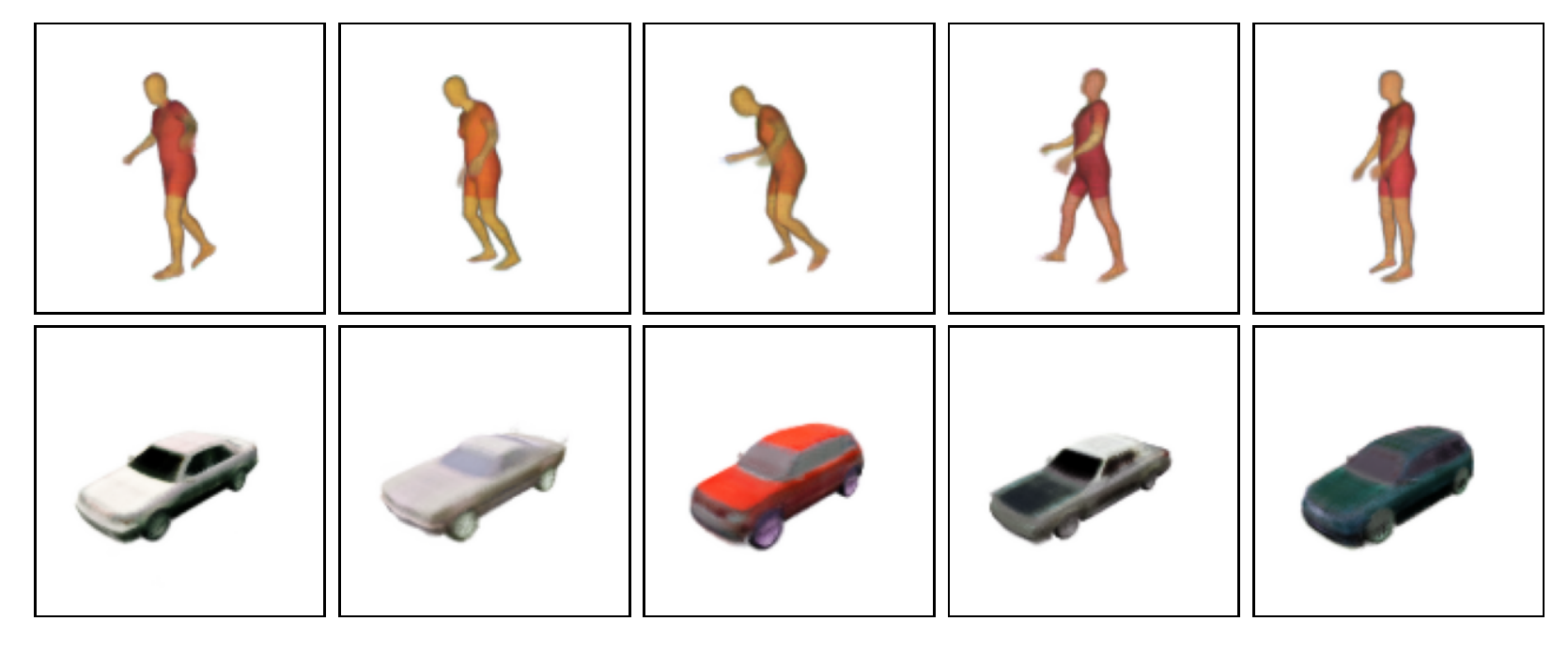}
    \caption{Unconditional GHUM and SRN Cars samples.}
    \label{fig:prior_samples}
\end{figure}

Our unconditional samples look realistic for both GHUM and SRN Cars (\cref{fig:prior_samples}; more in the supplementary material).
SRN Cars samples are on par with Functa in FID when it is trained using a denoising diffusion probabilistic model (DDPM) prior \citep{ho2020denoising} (\cref{tab:prior_sample_fid}).
Both ProbNeRF and Functa have a worse FID than $\pi$-GAN \citep{chan2021pi} which focuses only on generation and cannot be trivially extended to do uncertainty-aware inference.
We hypothesize that the gap between our retrained Functa model and the results reported by \cite{dupont2022data} are due to the lower expressivity of the RealNVP prior compared to a DDPM prior (although DDPMs are much more expensive to sample from and do posterior inference with).
\begin{table}[!t]
\begin{tabular}{@{}lllll@{}}
\toprule
    & ProbNeRF & Functa & Functa (DDPM) & $\pi$-GAN \\ \midrule
FID $\downarrow$ & 84.6     & 158.1            & 80.3          & 36.7   \\ \bottomrule
\end{tabular}
\caption{FID for SRN Cars prior samples.}
\label{tab:prior_sample_fid}
\end{table}

\subsection{Ablations}

\subsubsection{Exact Renderer}
\label{sec:exp-foam}

\begin{figure}[!h]
    \centering
    \includegraphics[width=\linewidth]{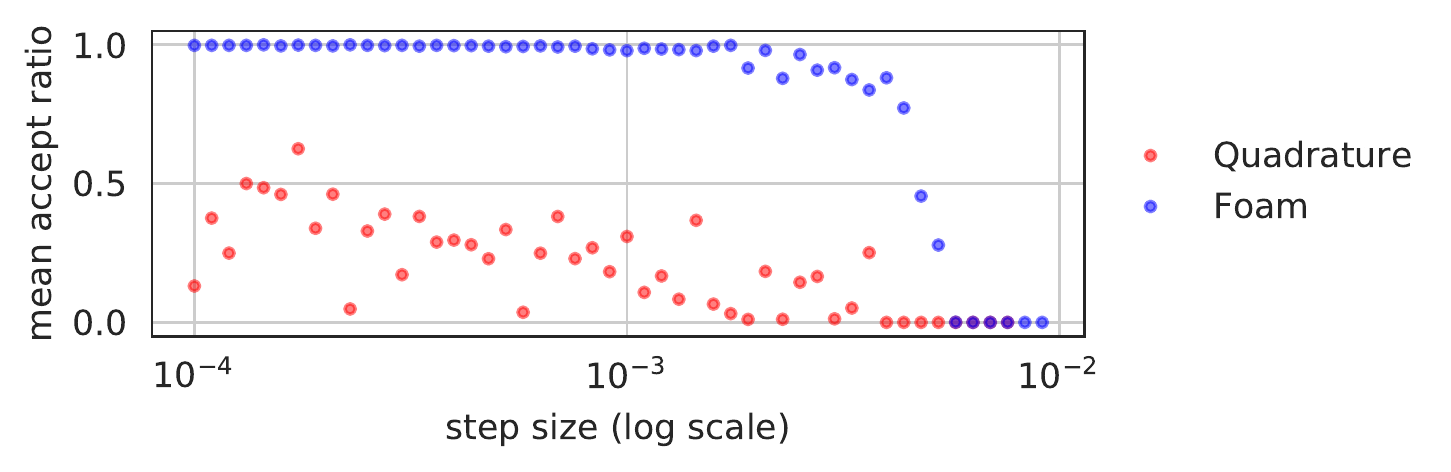}
    \vskip -0.1in
    \caption{HMC acceptance rates for quadrature and foam renderers.}
    \label{fig:foam}
\end{figure}

To demonstrate the difficulty of doing HMC with the approximate quadrature-based volumetric renderer \citep{mildenhall2020nerf} versus the exact foam renderer (\cref{sec:foam}), we first trained models using the quadrature- and foam-based renderers on SRN cars. We evaluate HMC on each of these models using the appropriate renderer as follows. We sample a fixed $\tilde z, \delta$ pair from the prior for each model and render a single view $y$ of the resulting NeRF; conditioned on this view, $\tilde z, \delta$ is a sample from the posterior $p(\tilde z, \delta\mid y, r)$. For each of a variety of step sizes, we then run 8 HMC chains with 10 leapfrog steps initialized from this sample and targeting $p(\tilde z, \delta\mid y, r)$ for 20 iterations, and report the average Metropolis acceptance rate across the chains on the last iteration. HMC sampling using the quadrature renderer suffers from low acceptance rates even with tiny step sizes, while the foam renderer yields high acceptance rates for small-enough step sizes (\cref{fig:foam}).

\subsubsection{Temperature Annealing}
\label{sec:exp-annealing}

\begin{figure}[!t]
    \centering
    \includegraphics[width=\linewidth]{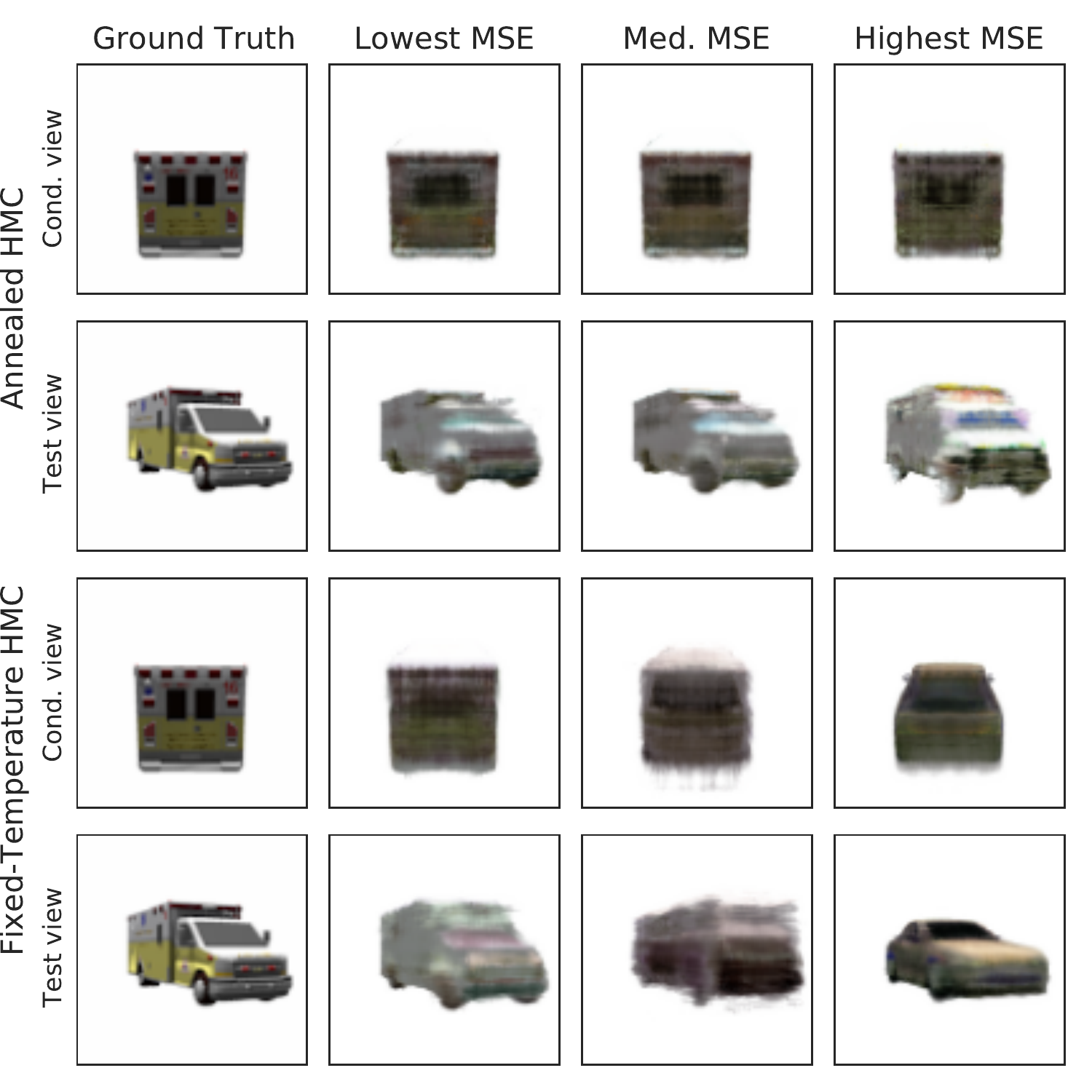}
    \caption{Renderings of samples generated by annealed and fixed-temperature HMC conditioned on an image of the rear of an ambulance (columns 2--4) and corresponding ground-truth images (column 1). Rows 1 and 3 show the conditioned-on view, rows 2 and 4 show a held-out view. Samples are selected to have the lowest (column 2), 4th-lowest (column 3), or highest (column 4) squared-error reconstruction of the conditioned-on view. Samples from all 8 chains are in the supplement.}
    \label{fig:annealing}
\end{figure}

In this section we qualitatively demonstrate the value of our annealed-HMC strategy. Conditioning on the rear view of an ambulance, we ran HMC with annealing as described in \cref{sec:hmc}, and compare with HMC initialized in the same way but without annealing and with a fixed step size of 0.0005 (we found it necessary to use a lower step size for fixed-temperature HMC to ensure reasonable acceptance rates across all chains). We ran both methods with 8 parallel chains. \cref{fig:annealing} shows the last samples from each method.

The annealed-HMC procedure's samples are both more consistent and more faithful to the ground truth. This result is consistent with the hypothesis that the annealing procedure allows HMC to avoid low-mass modes of the posterior and focus on more-plausible explanations of the data.

\subsubsection{Inference Over Raw NeRF Weights}

In addition to running HMC on ProbNeRF targeting both the latent code and the raw NeRF weights, $p(\tilde z, \delta | y, r)$, we run HMC on ProbNeRF targeting only the latent code, $p(\tilde z | y, r)$.
Like in \cref{sec:posterior_inference_probnerf}, we obtain $L = 8$ samples by running $L$ independent chains.
We pick the 16 samples from each chain and render reconstructions from the conditioning view and $H = 4$ held-out views.

We show that performing inference over the raw NeRF weights significantly increases the quality (higher PSNR) and realism (lower FID) of the conditioned-on view reconstruction while not having negative effects on held-out view reconstruction performance (\cref{fig:ablation_evals}).
Further, when conditioning on high-information views, the quality and realism of held-out views improves relative to when we only infer the latent code.
This supports our hypothesis that adding raw NeRF weights as latent variables increases the support with a positive prior over the radiance fields which lets our system adapt to novel views given sufficiently informative observations.
Importantly, despite doing test-time inference in a model that is different from the model used during training does not harm performance on reconstruction of far away held-out views or prior sampling.
For GHUM, the discrepancy between ProbNeRF and latent-only ProbNeRF conditioned-on view reconstruction quality is not as large, which we attribute to (i) a smaller discrepancy between training and test distributions and (ii) both models being able to fit the training distribution well.

\section{RELATED WORK}

\textbf{Neural fields for novel-scene inference.}
While classic NeRFs \citep{mildenhall2020nerf} are only fit on single scenes, there have been many recent extensions allowing novel scene or novel object inference.
CodeNeRF \citep{jang2021codenerf} and LOLNeRF \citep{rebain2022lolnerf} condition neural fields by concatenating inputs with per-scene latent codes, and learn priors that are able to generate coherent 3D geometry.
PixelNeRF \citep{yu2021pixelnerf}, IBRNet \citep{wang2021ibrnet}, GRF \citep{trevithick2021grf}, and MVSNeRF \citep{chen2021mvsnerf} exploit the geometry of the conditioning view, known as ``local conditioning'', to inform novel-view reconstructions.
As noted by \citet{sajjadi2022scene}, local conditioning generalizes less well to views far from observations, and is more computationally expensive.
\citet{sajjadi2022scene} explore using attention-based mechanisms alongside a set-based latent representation which \citet{rebain2022attention} observes to be generally superior to concatenation or hypernetwork-based methods.
However, hypernetworks are shown to perform nearly as well as attention mechanisms, and attention is expensive, especially for iterative posterior inference \citep{kosiorek2021nerf}.
Scene representation networks \citep{sitzmann2019scene} also use hypernetwork conditioning, although on a different neural representation.
Similar to ShaRF \citep{rematas2021sharf}, we also find that test-time inference of NeRF weights alongside latent codes improves reconstructions, especially when input images are highly informative.

\textbf{Probabilistic neural scene representations.}
Like ProbNeRF, these methods go beyond single point estimates and can represent multiple plausible scenes consistent with the potentially low-information views.
Generative Query Networks \citep{eslami2018neural,rosenbaum2018learning} also train a VAE system but use a convolutional decoder which doesn't enforce multi-view consistency.
NeRF-VAE \citep{kosiorek2021nerf} also combines VAEs and NeRFs, but relies on latent-concatenation conditioning and amortized VI, resulting in low-diversity novel-view reconstructions.
Concurrently with our work, \citet{anonymous2023laser} extends NeRF-VAE to produce larger posterior diversity by using normalizing flows, attention, and set-based latent representations; we were unable to evaluate it at the time of submission.
GAUDI \citep{bautista2022gaudi} fits a diffusion-based conditional generative model that can sample diverse and plausible large-scale real world scenes given few observed views, although the conditioning mechanism must be trained from paired data, whereas ProbNeRF can condition on arbitrary sets of pixels and camera positions.
3DiM \citep{watson20223dim} is a diffusion-based image-to-image model that can synthesize diverse novel views of scenes, but it does not guarantee multi-view consistency.

\textbf{Other NeRF generative models.}
\citet{schwarz2020graf,niemeyer2020giraffe,chan2021pi} train NeRF generative models using a discriminator loss from generative adversarial nets (GANs) \citep{goodfellow2014generative} which produce high-quality, diverse samples.
However, GAN-style training often results in models that struggle at reconstruction \citep{wu2017on}.
DreamFusion \citep{poole2022dreamfusion} generates impressive NeRF scenes given a text prompt, but also does not focus on inference from images.

\textbf{Exact rendering.} DIVeR \citep{wu2021realtime} introduces a deterministic and exact renderer based on integrating trilinearly interpolated features exactly on a voxel grid.
This requires four times as many function evaluations or table lookups per intersected voxel as the MobileNeRF strategy we adapt \citep{chen2022mobilenerf}.

\textbf{Probabilistic programming for computer vision.} Although several probabilistic programming approaches to computer vision use test-time Monte Carlo  inference \citep{mansinghka2013approximate, le2017inference, kulkarni2015picture, gothoskar20213dp3}, they mainly focus on finding one probable scene interpretation per 2D image
(though \citet{mansinghka2013approximate} demonstrate limited domain-specific uncertainty reporting on a restricted class of road scenes).
In contrast, ProbNeRF characterizes shape and appearance uncertainty for open-ended classes of 3D objects.

\section{DISCUSSION}

Given a single low-information view of a novel object, ProbNeRF can not only produce reasonable point estimates of that object's shape and appearance, but can also guess what range of shapes and appearances are consistent with the available data. 
Making these sorts of diverse (but coherent) guesses about unseen features of objects is a fundamental problem in in vision.
ProbNeRF shows that it is possible to simultaneously achieve high-fidelity reconstruction and robust characterization of uncertainty within the NeRF framework. One next step for future research could be to quantify tradeoffs between model fidelity, inference efficiency, and uncertainty characterization, to support variations on ProbNeRF suitable for real-time applications in robotics. More broadly, we hope ProbNeRF encourages more research at the interface of Bayesian inference, 3D graphics, and computer vision, enabling computer vision systems that entertain diverse hypotheses about the world.

\acknowledgments{
  Thanks to Katie Colins, Varun Jampani, Adam Kosiorek, Despoina Paschalidou, and Sharad Vikram for their helpful comments, and to Alex Alemi, Kevin Murphy, and Andrea Tagliasacchi for their timely and helpful feedback on the manuscript. 
}

\bibliography{main}

\clearpage
\begin{appendices}

\onecolumn

\crefalias{section}{appsection}

\section{Additional figures}

\begin{figure*}[!htb]
    \centering
    \includegraphics[width=\linewidth]{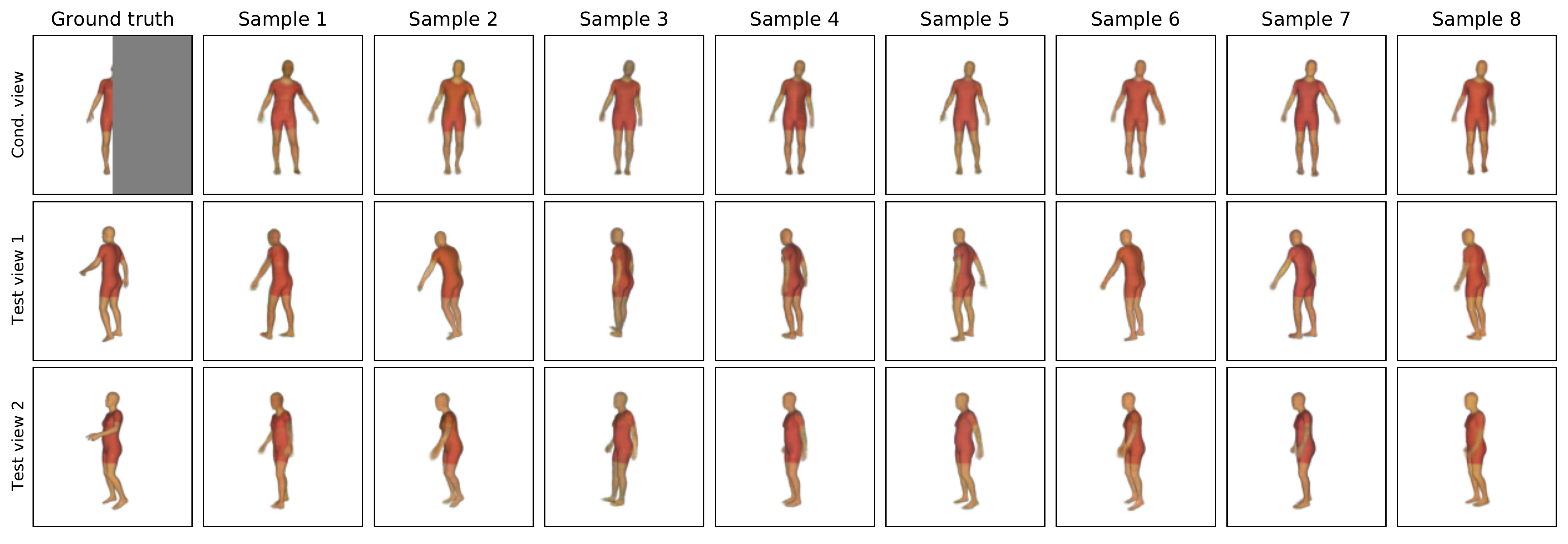}
    \caption{Uncurated GHUM HMC samples.}
\label{fig:full_hmc_ghum}
\end{figure*}

\begin{figure*}[!htb]
    \centering
    \includegraphics[width=\linewidth]{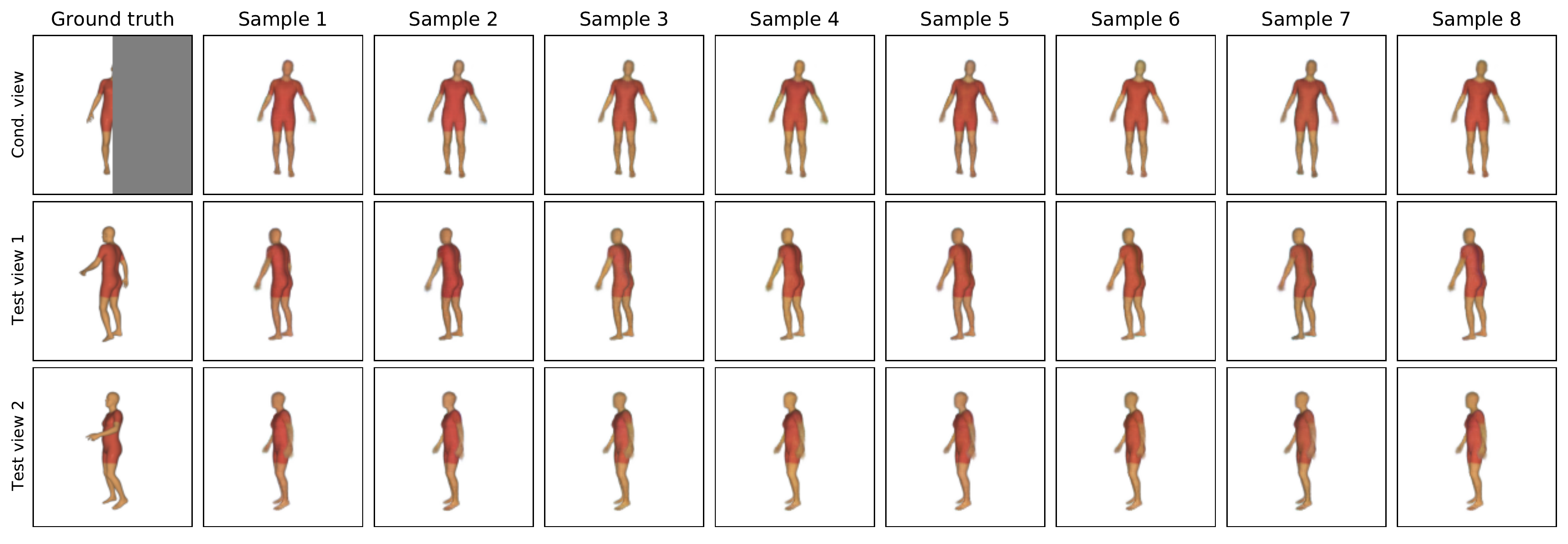}
    \caption{Uncurated GHUM VI samples.}
\label{fig:full_vi_ghum}
\end{figure*}

\begin{figure*}[!htb]
    \centering
    \includegraphics[width=\linewidth]{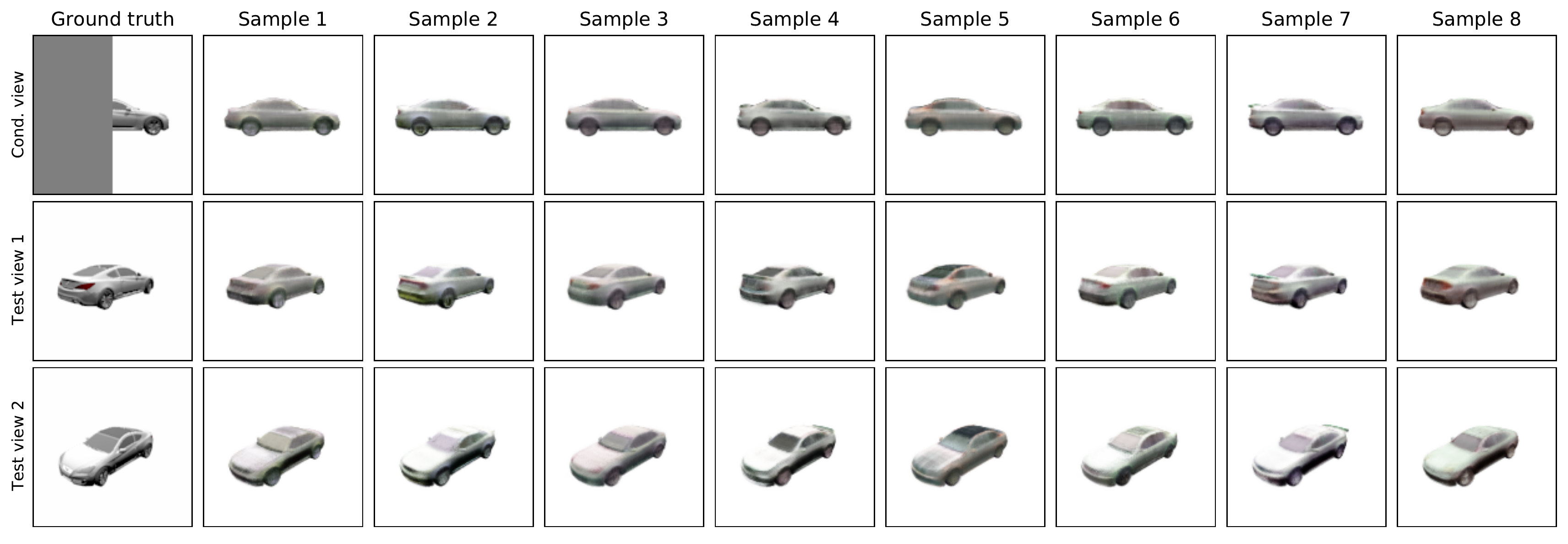}
    \caption{Uncurated SRN Cars HMC Samples}
\label{fig:full_hmc_srn}
\end{figure*}

\begin{figure*}[!htb]
    \centering
    \includegraphics[width=\linewidth]{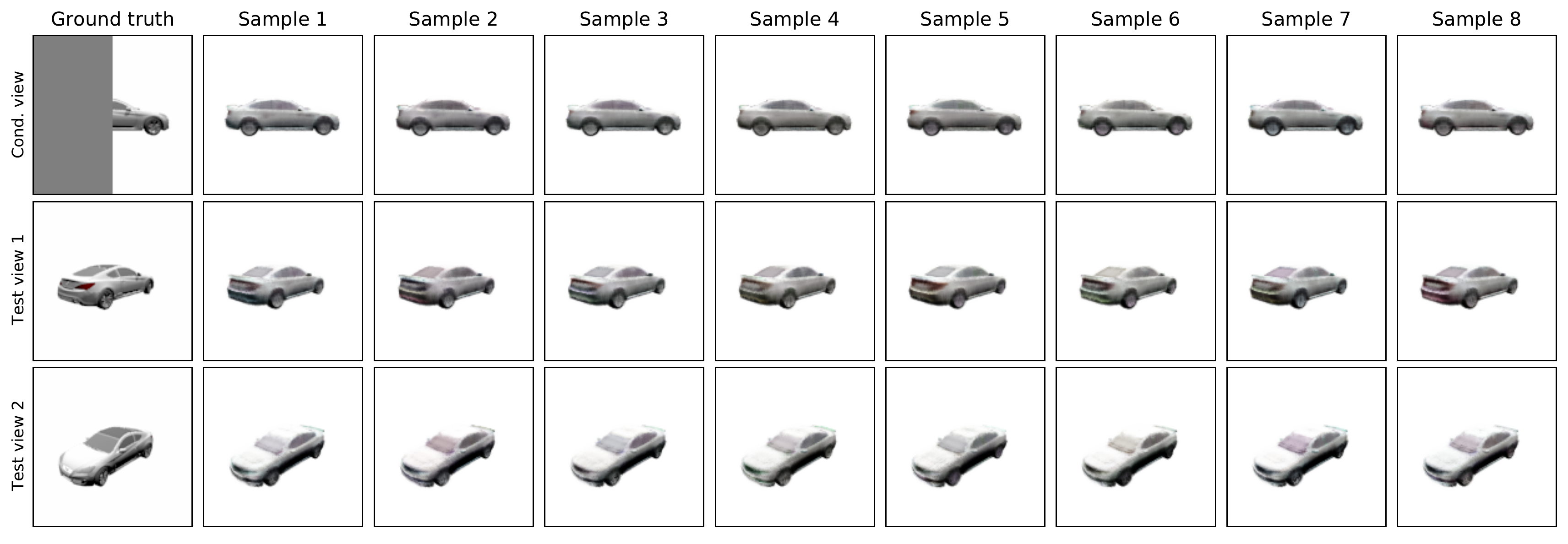}
    \caption{Uncurated SRN Cars VI samples.}
\label{fig:full_vi_srn}
\end{figure*}

\begin{figure*}[!htb]
    \centering
    \includegraphics[width=\linewidth]{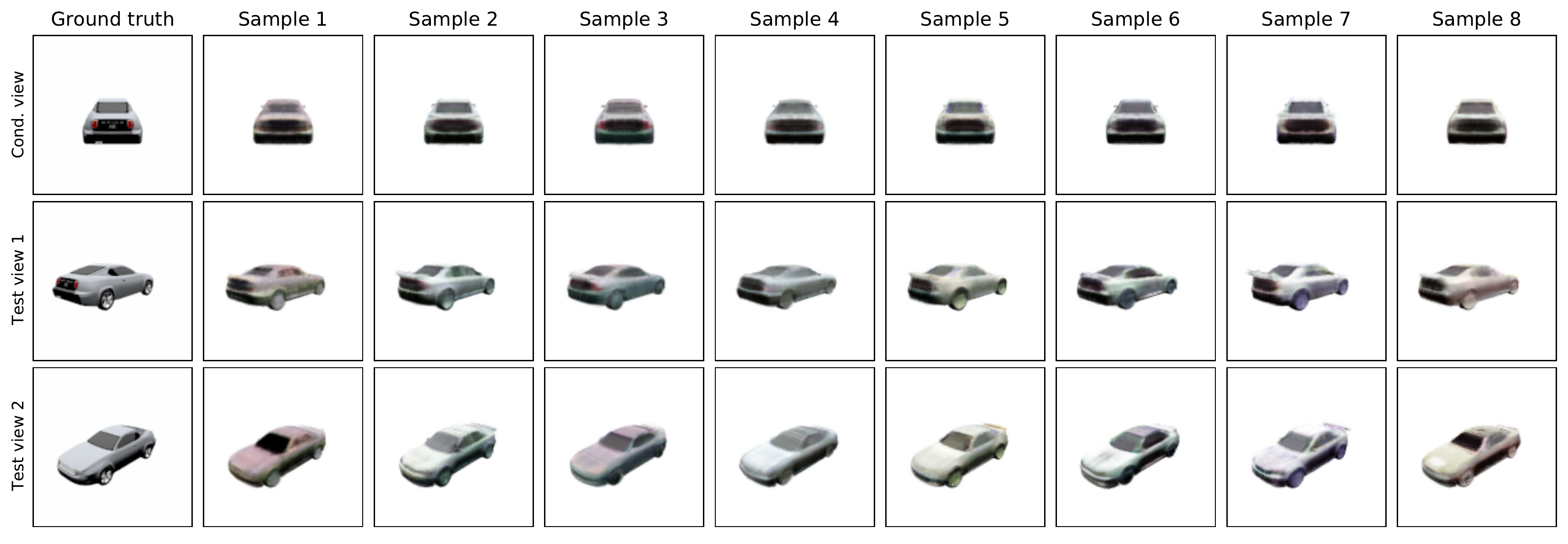}
    \caption{Full-view grey SRN Cars HMC samples.}
\label{fig:full_view_hmc_srn}
\end{figure*}

\begin{figure*}[!htb]
    \centering
    \includegraphics[width=\linewidth]{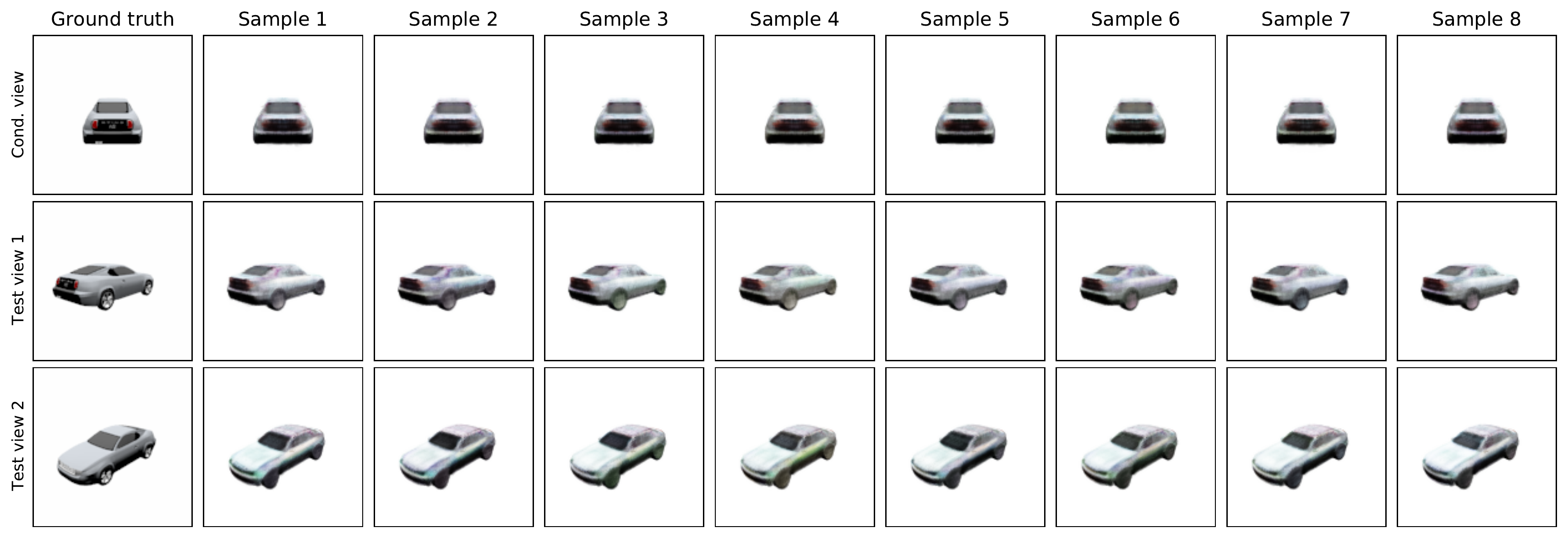}
    \caption{Uncurated full-view grey SRN Cars VI samples.}
\label{fig:full_view_vi_srn}
\end{figure*}

\begin{figure*}[!htb]
    \centering
    \includegraphics[width=\linewidth]{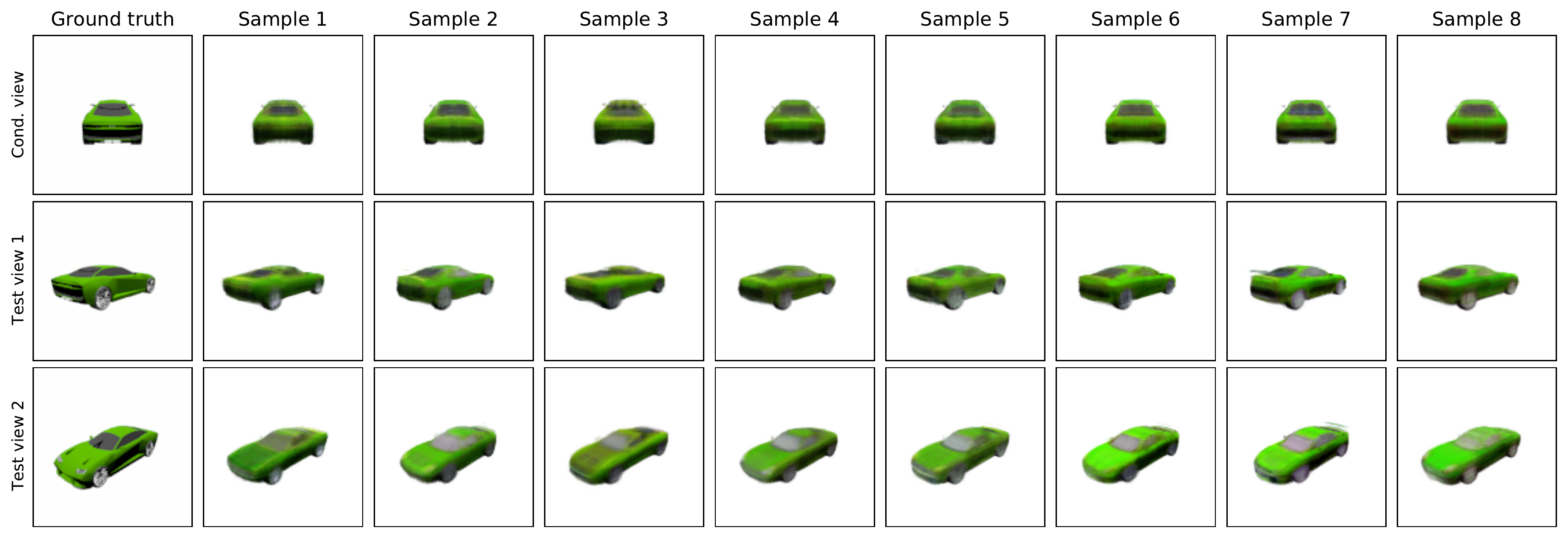}
    \caption{Uncurated full-view green SRN Cars HMC samples.}
\label{fig:full_view_hmc_srn_green}
\end{figure*}

\begin{figure*}[!htb]
    \centering
    \includegraphics[width=\linewidth]{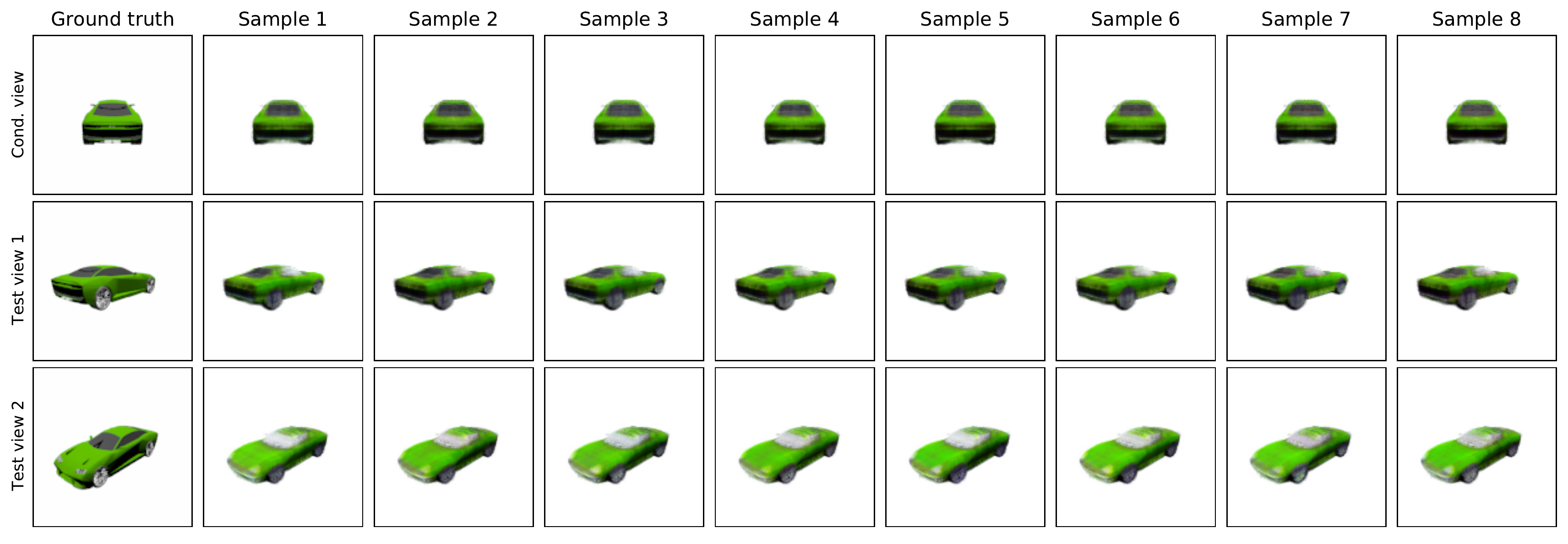}
    \caption{Uncurated full-view green SRN Cars VI samples.}
\label{fig:full_view_vi_srn_green}
\end{figure*}

Figures \ref{fig:full_hmc_ghum}, \ref{fig:full_vi_ghum}, \ref{fig:full_hmc_srn}, and \ref{fig:full_vi_srn} show eight uncurated samples from HMC and VI for GHUM and SRN Cars (cf. \cref{fig:ghum_and_srncars_samples_and_variance}). Figures \ref{fig:full_view_hmc_srn}, \ref{fig:full_view_vi_srn}, \ref{fig:full_view_hmc_srn_green}, and \ref{fig:full_view_vi_srn_green} show eight uncurated HMC and VI samples generated by conditioning on uncropped single-view images of two SRN cars. Both HMC and VI produce models that are realistic and consistent with the observed data, but samples from the variational distribution obtained by VI are all essentially the same, while HMC produces samples with noticeable diversity in pose (for GHUM) and shape and color (for SRN Cars).

\begin{figure*}
    \centering
    \includegraphics[width=\linewidth]{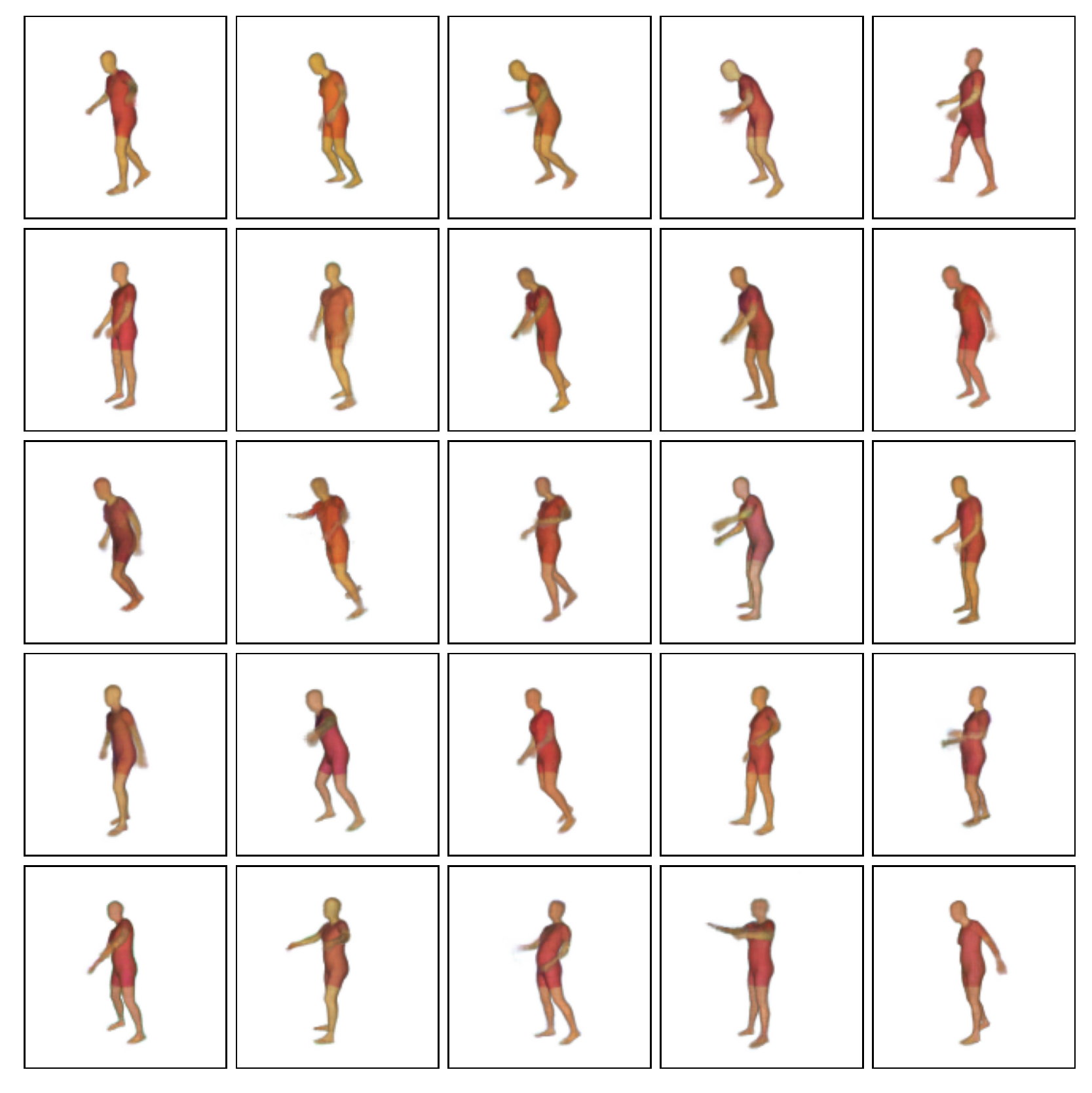}
    \caption{GHUM unconditional samples.}
    \label{fig:ghum_prior_samples}
\end{figure*}

\begin{figure*}
    \centering
    \includegraphics[width=\linewidth]{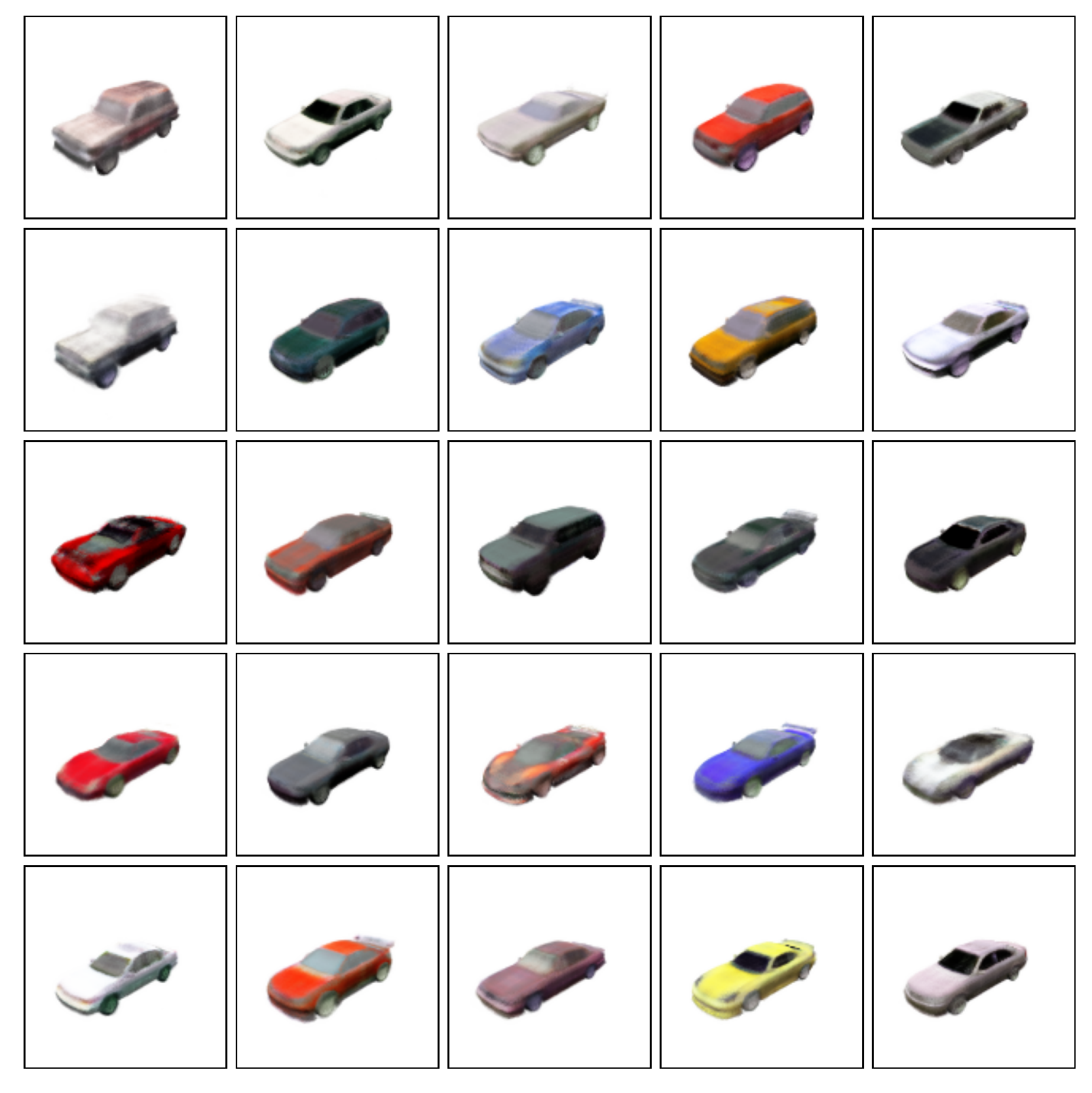}
    \caption{SRN Cars unconditional samples.}
    \label{fig:srncars_prior_samples}
\end{figure*}

Figures \ref{fig:ghum_prior_samples} and \ref{fig:srncars_prior_samples} show uncurated unconditional samples from the trained models.

\begin{figure*}
    \centering
    \includegraphics[width=\linewidth]{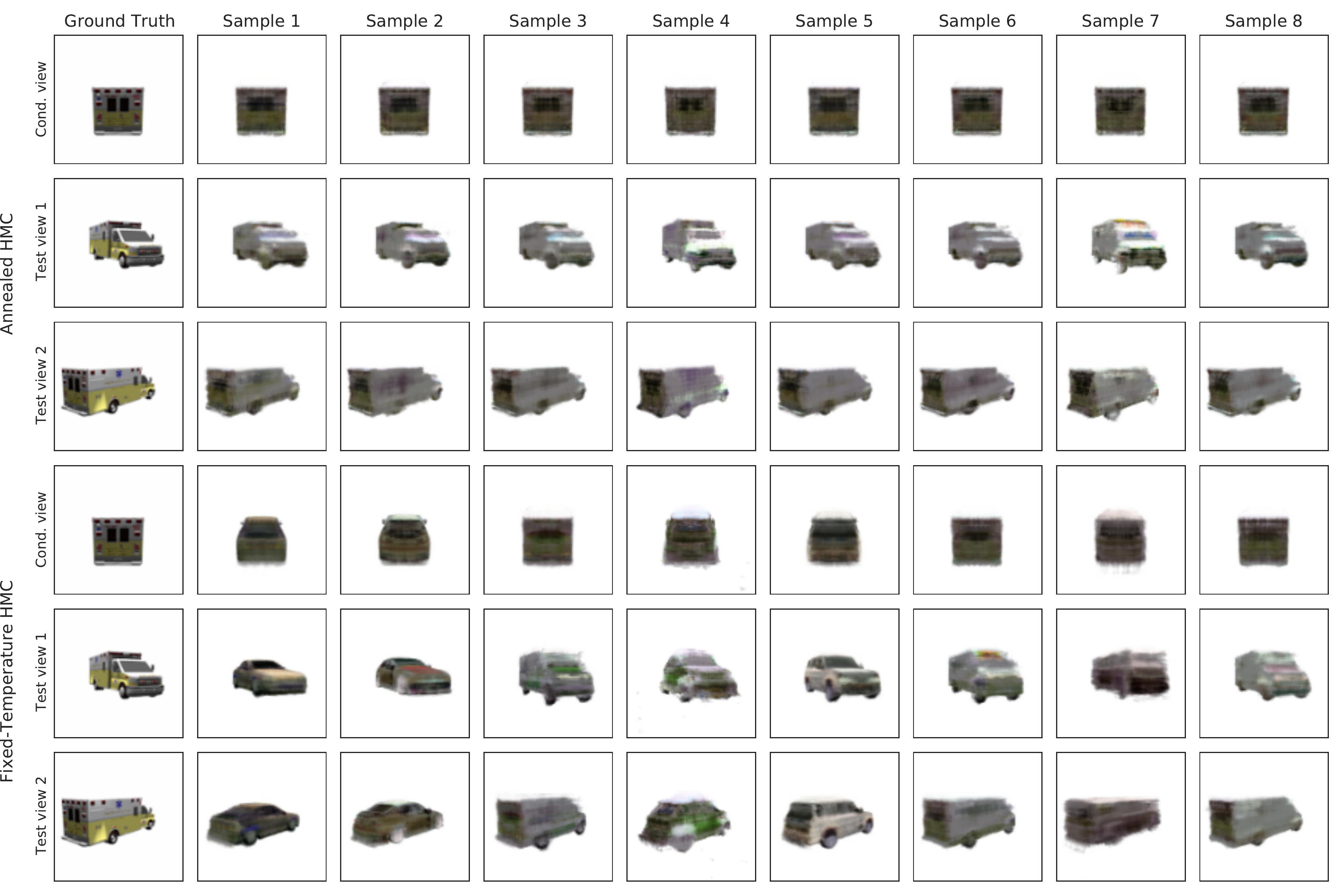}
    \caption{Uncurated annealed and fixed-temperature HMC samples.}
    \label{fig:annealing_full}
\end{figure*}

\cref{fig:annealing_full} shows uncurated samples conditioned on the rear view of an ambulance (cf. \cref{fig:annealing}) generated using HMC with temperature annealing and with a fixed temperature. Annealed HMC consistently finds solutions that are consistent with the conditioned-on view; fixed-temperature HMC does not.

\section{Architecture details}

\begin{figure*}[!htb]
    \centering
    \includegraphics[width=\linewidth]{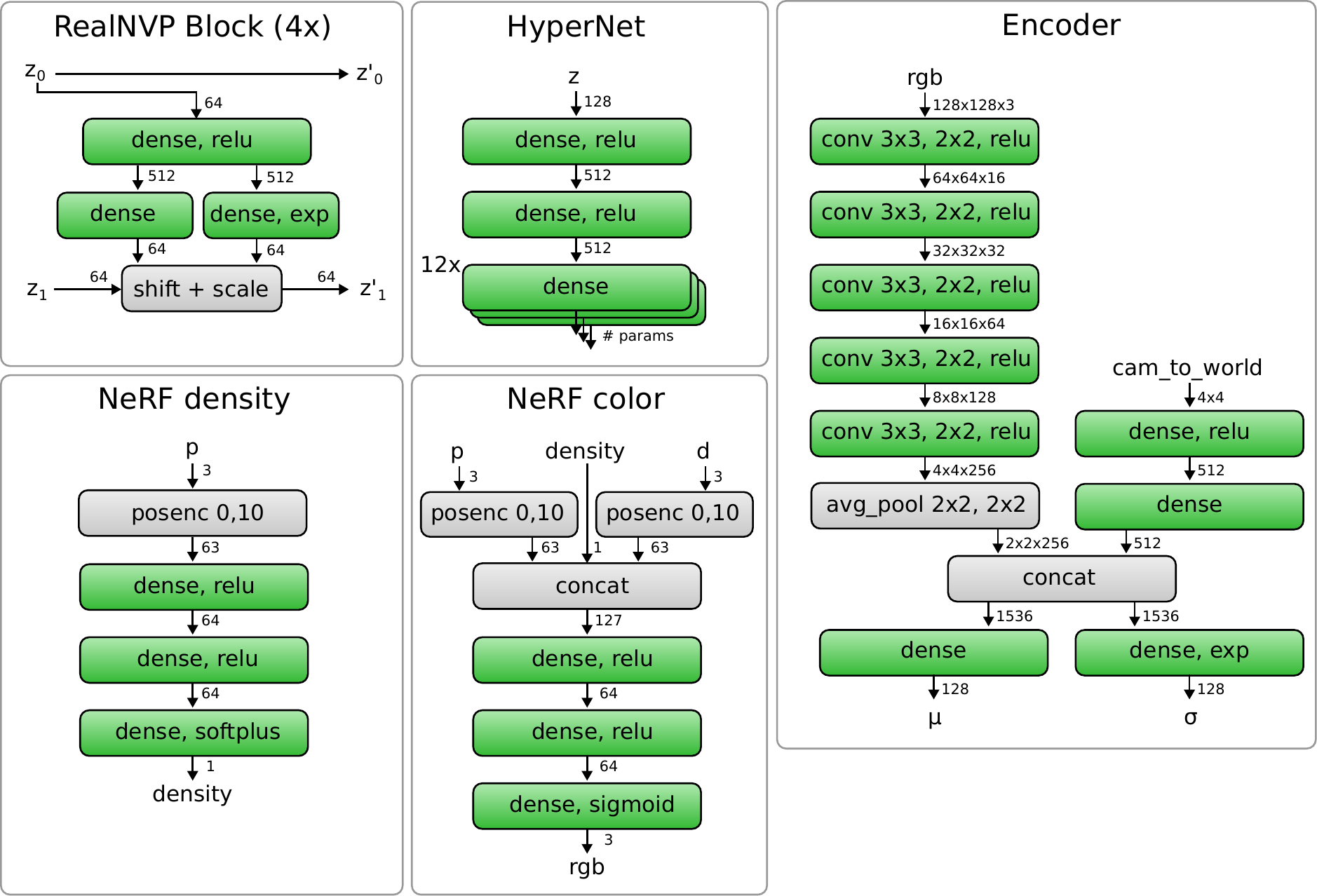}
    \caption{Neural Nets used in ProbNerf}
    \label{fig:net_diagram}
\end{figure*}

In this section we briefly describe the architectural details of ProbNerf neural nets. \cref{fig:net_diagram} contains the architectures used in this work.

\textbf{HyperNet} $h(z; \theta)$ is a simple MLP with two shared hidden layers, followed by a learnable linear projection and reshape operations to produce the parameters of the two NeRF networks.

\textbf{RealNVP} $m(\tilde z; \zeta)$ consists of 4 RealNVP blocks which act on a latent vector split into 2 parts (designated as $z_0$ and $z_1$ in the diagram). The split sense is reversed between the RealNVP blocks.

\textbf{NeRF} The NeRF is split into two sub-networks, one for density and one for color. The input position $p$ and ray direction $d$ are encoded using a 10th order sinusoidal positional encoding. For a scalar component of the input vector $x_i$ we produce a feature:
\begin{align}
    f_{i} = \{ \sin(2^j \pi x_i + 0.5 k) | j\in[0, 10), k\in[0, 1] \}.
\end{align}
We flatten and concatenate this array with the original input value to produce a 21 element feature vector for each $x_i$. To convert output density $\sigma\in\mathbb{R}^+$ to $\alpha\in[0, 1]$ we squash it as $\alpha = 1 - \exp(-\sigma / 128)$, where 128 is the grid size.

\textbf{Encoder} Each potential of the variational posterior is modeled as a diagonal covariance Gaussian with mean $\mu$ and scale $\sigma$ computed via a CNN.

\section{Equivalence of linear latent-shift modulations and latent concatenation}

The linear shift-only modulations used by \citet{dupont2022data} work as follows for an MLP: given a latent vector $z$, for each layer's output pre-nonlinearity activations $a^{(i)}$ (treating the input as an activation vector $a^{(0)}$), add a shift vector $s^{(i)} = V^{(i)}z$ that is a linear function of $z$ to get $a'^{(i)} = a^{(i)} + s^{(i)}$, and propagate $a'$ forward through the network instead of $a$.

The same effect can be achieved by concatenating $z$ to the activations $a'^{(i)}$ at each level $i$ of the network. The resulting computation is
\begin{equation}
\begin{split}
a'^{(i)} &= W^{(i)}(\sigma(a'^{(i-1)})^\top, z^\top)^\top + b^{(i)}
\\&\triangleq \tilde W^{(i)}\sigma(a'^{(i-1)}) + b^{(i)} + V^{(i)}z
\\&\equiv a^{(i)} + s^{(i)},
\end{split}
\end{equation}
where $W^{(i)}$ denotes the weight matrix at layer $i$, $b^{(i)}$ denotes the biases at layer $i$, $\sigma$ denotes a nonlinear activation function, and we define $\tilde W^{(i)}$ and $V^{(i)}$ to be the submatrices of $W^{(i)}$ that are multiplied by the previous layer's activations $\sigma(a'(^{i-1}))$ and the concatenated latents $z$ respectively.

\end{appendices}

\end{document}